\documentclass[journal]{ieeetran}
\usepackage[pdftex]{graphicx}
\graphicspath{{../pdf/}{../jpeg/}}
\DeclareGraphicsExtensions{.pdf,.jpeg,.png}
\usepackage[T1]{fontenc} % optional enhanced font encoding
\usepackage{amsmath, amsfonts, amssymb, amsthm}
\usepackage[cmintegrals]{newtxmath}
\usepackage{bm}
\usepackage{array}
\usepackage{url}
\usepackage{subcaption}
\usepackage{booktabs}
\usepackage{multirow}
\usepackage{multicol}
\usepackage{xcolor}

\newcommand{\red}[1]{\textcolor{black}{#1}}

\begin{document}
% The paper headers
% \markboth{Vol.~1, No.~3, May~2023}{0000000}

% article subject line 
% \IEEELSENSarticlesubject{Sensor Applications}

% paper title
\title{Object Depth and Size Estimation using Stereo-vision and Integration with SLAM}

\author{
\IEEEauthorblockN{
Layth~Hamad \IEEEauthorrefmark{1},~\IEEEmembership{Member,~IEEE,}
Muhammad Asif~Khan \IEEEauthorrefmark{1},~\IEEEmembership{Senior Member, IEEE,} 
and~Amr~Mohamed \IEEEauthorrefmark{2},~\IEEEmembership{Senior~Member,~IEEE}}

\IEEEauthorblockA{
\IEEEauthorrefmark{1}Qatar Mobility Innovations Center, Qatar University, Doha, Qatar.\\
\IEEEauthorrefmark{2}Qatar University, Doha, Qatar.}
}

\maketitle
% \IEEEtitleabstractindextext{%
\begin{abstract}
Autonomous robots use simultaneous localization and mapping (SLAM) for efficient and safe navigation in various environments. LiDAR sensors are integral in these systems for object identification and localization. \red{However, LiDAR systems though effective in detecting solid objects (e.g., trash bin, bottle, etc.), encounter limitations in identifying semitransparent or non-tangible objects (e.g., fire, smoke, steam, etc.) due to poor reflecting characteristics. Additionally, LiDAR also fails to detect features such as navigation signs and often struggles to detect certain hazardous materials that lack a distinct surface for effective laser reflection. In this paper, we propose a highly accurate stereo-vision approach to complement LiDAR in autonomous robots. The system employs advanced stereo vision-based object detection to detect both tangible and non-tangible objects and then uses simple machine learning to precisely estimate the depth and size of the object.} The depth and size information is then integrated into the SLAM process to enhance the robot’s navigation capabilities in complex environments. Our evaluation, conducted on an autonomous robot equipped with LiDAR and stereo-vision systems demonstrates high accuracy in the estimation of an object's depth and size. A video illustration of the proposed scheme is available at: \url{https://www.youtube.com/watch?v=nusI6tA9eSk}.

\end{abstract}


% If you want to put a publisher's ID mark on the page you can do it like
% this:
% \IEEEpubid{1949-307X \copyright\ 2023 IEEE. Personal use is permitted, but republication/redistribution requires IEEE permission.\\
% See \url{http://www.ieee.org/publications\_standards/publications/rights/index.html} for more information.}
% Remember, if you use this you must call \IEEEpubidadjcol in the second
% column for its text to clear the IEEEpubid mark.

% make the title area
% \maketitle

\section{Introduction}
% \IEEEPARstart{T}{his} 
Autonomous mobile robots (AMRs) have witnessed a transformative revolution in various applications, from manufacturing \cite{robot_industry2021, robot_manuf2019} to healthcare \cite{Mukherjee2022}, agriculture \cite{Ball2017} to logistics \cite{Shangguan2021}, and beyond. A fundamental requirement for these autonomous systems is the ability to perceive and understand their environment accurately. Achieving this understanding is pivotal for safe and efficient navigation, obstacle avoidance, and task execution.
AMRs are typically equipped with various sensors that enable them to perceive their environment and optimize their routes dynamically. This is achieved using simultaneous localization and mapping (SLAM). There are various sensor modalities that can be used in SLAM e.g., LiDAR, vision-based, inertial sensors, etc. LiDAR-based SLAM uses laser range finders to capture data and accurately represent the geometry of the environment. Vision-based SLAM is becoming more popular due to the availability of more accurate off-the-shelf object detection models.
However, LiDAR has some intrinsic limitations in detecting low-lying objects, materials with low reflectance, and small-sized obstacles, particularly in complex and dynamic environments. These limitations inherent in LiDAR technology have prompted the search for alternative solutions.
Vision-based systems are getting popular mainly due to low cost, ease of deployment, and the availability and high performance of off-the-shelf object detection models \cite{Strbac2020, Kazerouni2023, visual_slam}.
\par
This paper proposes a framework using deep learning for obstacle detection and stereo vision for estimating the depth and size of the detected objects in the real world to enhance the capabilities of autonomous robots. The proposed stereo-vision image mimics the human visual system for environmental perception. The system uses two horizontally spaced cameras installed on a robot to simultaneously capture images of the environment and then use deep learning-based object detection to detect potential obstacles in its navigation path. As object detection only predicts the category of the object and its location inside the image, the dual images are further processed to find disparities in object locations. The object location (inside the images) and the relative disparities in their locations in the two images are then used to estimate the object's depth and dimensions in the real world. The proposed framework can be adopted in several use cases for mobile robot navigation and can also be extended in other applications such as drone localization \cite{khan2022detection}.

The main contribution of this paper is as follows:
\begin{itemize}
\item We propose an intuitive method based on stereo-vision for object depth and size estimation from the output of the object detection models. This works in two stages. First, the object depth is estimated from the bounding box coordinates of the detected objects in corresponding images from two cameras. Then the estimated depth of the object and the bounding box width and height are used to estimate the object size in the real world.
\item The depth and size estimations are then integrated into the SLAM system to translate the depth and size information into the robotic environment for real-time obstacle avoidance and navigation.
\end{itemize}

%===========================================
\section{Related Work}
%===========================================
%% 72, 73, 74, 
% Traditional block matching algorithms, like Sum of Absolute
% Differences (SAD) Tippetts [78] and Sum of Squared Differences (SSD) [79], calculate
% matching costs based on fixed windows in the right images. Semi-Global Matching (SGM) takes this approach further by considering information through wise optimization,
% resulting in improved depth maps by minimizing energy functions[80].
There have been different approaches to measuring object depth and dimensions using computer vision. For instance, authors in \cite{chen2019} proposed a distance estimation method using object detection. The method uses YOLOv3 \cite{yolov3} to detect the bounding boxes first and then compute the histogram from the disparity maps. However, the dataset used in this study does not have real-world distance annotations, and the authors could not verify the method's accuracy.
Authors in \cite{Kazerouni2023} implemented distance estimation using deep learning. Objects of interest are detected in a scene and the corresponding bounding box outputs are then cropped using the segmentation mask. Then the depth map is estimated using SGDepth \cite{SGDepth2020} for background removal. SGDepth estimates the depth information from a single image obtained from segmentation maps and does not require depth labels. The output of SGDepth (depth map) is processed for noise removal and is passed through a pooling layer to compute distances of regions of interest. Different from \cite{Kazerouni2023}, our proposed method is embarrassingly simple, yet highly intuitive and accurate in computing the size of the objects and the distance to the object, thus highly effective in real-time robot navigation systems.
Authors in \cite{Strbac2020} proposed a multi-camera object distance estimation using object detection and stereo-vision. This method uses object detection to detect object boundaries, find matching objects, and then use a mathematical expression to compute the distance between the object and the camera. However, the analytical model proposed in this work could produce accurate results only on a portion of data samples and heavily underperforms in many cases. A comparison of our proposed work against \cite{Strbac2020} has been presented in Section \ref{sec:results}. To the best of our knowledge, our proposed method inspired by human visual perception is a novel scheme that is highly intuitive, surprisingly simple to implement, and accurate.

%===========================================
\section{Proposed Framework}
%===========================================

\begin{figure*}[htbp]
    \centering
    \includegraphics[width=0.9\textwidth]{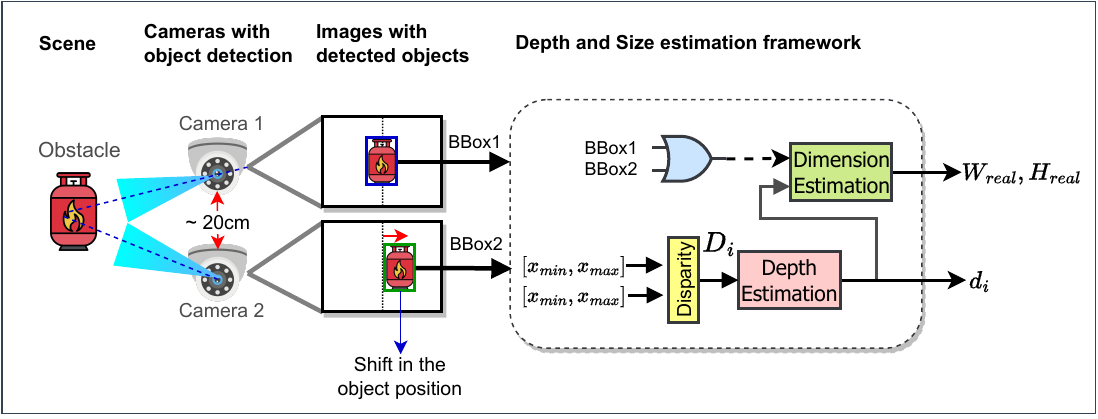}
    \caption{The proposed framework for estimating object depth (distance) and dimensions (size in the real world).}
    \label{fig:prop_sch}
\end{figure*}

The proposed stereo vision system used in the mobile robot is comprised of two horizontally oriented cameras installed on the left and right side of the top of the robot mimicking human eyes. The cameras are apart from each other at a distance of 20 cm. Both cameras concurrently capture images of a scene. The images are processed by an object detection model (i.e., YOLOv8) to detect objects of interest using the bounding box detection method. The bounding boxes are then analyzed to extract the depth (distance of the object from the camera) and dimension (size of the object in width and height) of the detected object in the real world. Fig. \ref{fig:prop_sch} illustrates the proposed framework.
The system operates in three distinct stages to determine (a) obstacle detection, (b) depth (distance) estimation, and (c) dimension (size ) estimation. The depth and size estimations are then used in the SLAM algorithm for obstacle avoidance and safe navigation of the robot.
%=========================================================
\subsection{Obstacle Detection}
Obstacle detection involves training the YOLOv8 object detection model and post-processing the model outputs. The object detection model is trained on a set of objects of interest. At inference time, the model returns the predictions comprising the predicted class and bounding box coordinates of the object (in the image). The prediction outputs on both images of the same scene (from the two cameras) are stored separately as predictions $A$ and $B$ as arrays. The output arrays $A$ and $B$ are then sorted such that the predictions in $A$ and $B$ at the same index correspond to the the same object. This is achieved by comparing the horizontal spacing (in pixels) between the two bounding boxes in respective images.
%=========================================================
\begin{figure}[!h]
    \centering
    \includegraphics[width=0.6\columnwidth]{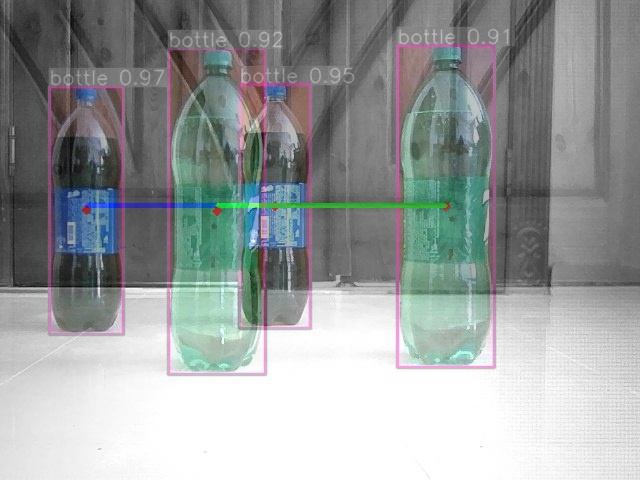}
    \caption{Stereo vision image with object detection from the two cameras of the robot showing the disparity of the objects.}
    \label{fig:blended_img}
\end{figure}

\subsection{Depth Estimation}
Depth estimation (horizontal distance between the object and the robot) is crucial for collision avoidance and safe navigation. To estimate the depth information, we exploit the post-processed bounding box outputs to compute the horizontal shift or disparity for each of the four points by determining their pixel offsets between the left and right camera images using Eq. \ref{eq:disparity}.
Fig. \ref{fig:blended_img} shows the relative disparities in two detected objects in dual images captured using the robot.
The disparity offset between the corresponding bounding boxes for the left and right camera images is calculated as follows:
\begin{equation} \label{eq:disparity}
D_i = \frac{1}{4} \left( \Delta x_{tl,i} + \Delta x_{tr,i} + \Delta x_{bl,i} + \Delta x_{br,i} \right)
\end{equation}

where \( \Delta x_{k,i} \) denotes the horizontal pixel offset of point \( k \) (e.g., \( tl, tr, bl, br \)) between the left and right images for the $i^th$ bounding box pair. Thus, each pair of the bounding boxes yields a unique disparity value \( D_i \), which captures the depth estimation localized to that specific object. The disparity values obtained from the stereo vision system are used as the primary input to a machine learning model to predict the actual depth.
%=========================================================
\subsection{Dimension Estimation}
Objects with varying real-world dimensions can manifest similar bounding box sizes within an image, primarily due to distance variations (perspective distortion). However, if the real-world distance between the camera and the object is known, the actual dimensions of the objects can be determined. Specifically, we leveraged the estimated depth of the object in conjunction with the bounding box size (in pixels), to deduce the actual dimensions of the object (in cm). We used polynomial regression of degree-k for this purpose. The model used depth and pixel dimensions as features to map them to the corresponding real-world dimensions (actual width and height of the object).

%=========================================================
\subsection{Integration with SLAM}
To exploit the estimated depth and size of the detected object, a pragmatic approach is proposed to integrate the processed object data within the SLAM system. Intuitively, due to the rectangular shapes of the bounding boxes (and the corresponding real-world size of the predicted objects), we opt to represent detected objects as cylindrical obstacles in the SLAM system. The cylinder's height corresponds to the estimated object height, while the diameter is set equal to the object's width. The relative $(x,y)$ coordinates serve to position the cylinder in alignment with the actual object's location with respect to the robot.
\par
For seamless assimilation into the Robot Operating System (ROS) ecosystem, we developed a module that emulates a point cloud ROS sensor. This virtual sensor continuously publishes the point cloud data of the constructed cylindrical obstacle, representing the detected object. Configured within ROS, this virtual sensor maintains a static transformation with respect to the robot. Consequently, the disseminated point cloud data effectively symbolizes the obstacles to be evaded, and this information is promptly integrated into both the global and local cost maps of the ROS navigation stack, ensuring accurate obstacle avoidance during robot navigation. The cylindrical representation of detected objects is then translated to a point cloud format by decomposing the surface of each cylinder into a dense collection of discrete points.

%===========================================
\section{Experiments and Results} \label{sec:results}
%===========================================

\subsection{Experiments}
% Detection
\red{We first trained an object detection model on a small dataset of objects of interest. YOLOv8 is used for object detection due to its higher accuracy over previous YOLO models \cite{Terven2023}. Then, a comprehensive dataset of paired images of different obstacles (e.g., bottle, book, jug, glasses, scissor, etc.) was curated comprising pairs of known disparities and their corresponding real-world depths and dimensions to train supervised learning models with high accuracy. Some example objects are shown in Fig. \ref{fig:slam_intgration}a}. 
% Depth
Given the intrinsic nonlinear relationship between disparity and depth, the model is designed to approximate this relationship via a polynomial fitting of degree 5. The choice of a quintic polynomial is pivotal, offering a balance between capturing the complex relationship and avoiding overfitting. 
In XGBoost \cite{XGBoost_2016}, the selection of polynomial's degree is crucial to balance learning and generalization. We trained several models and a polynomial of degree 5 was found to be a good fit. Furthermore, k-fold cross-validation was used with $k=5$ to prevent overfitting.
% Dimensions

\subsection{Evaluation and Results}
\begin{figure*}[!h]
    \centering
        \subcaptionbox{BBox disparity between stereo images.}
        {\includegraphics[width=0.28\textwidth]{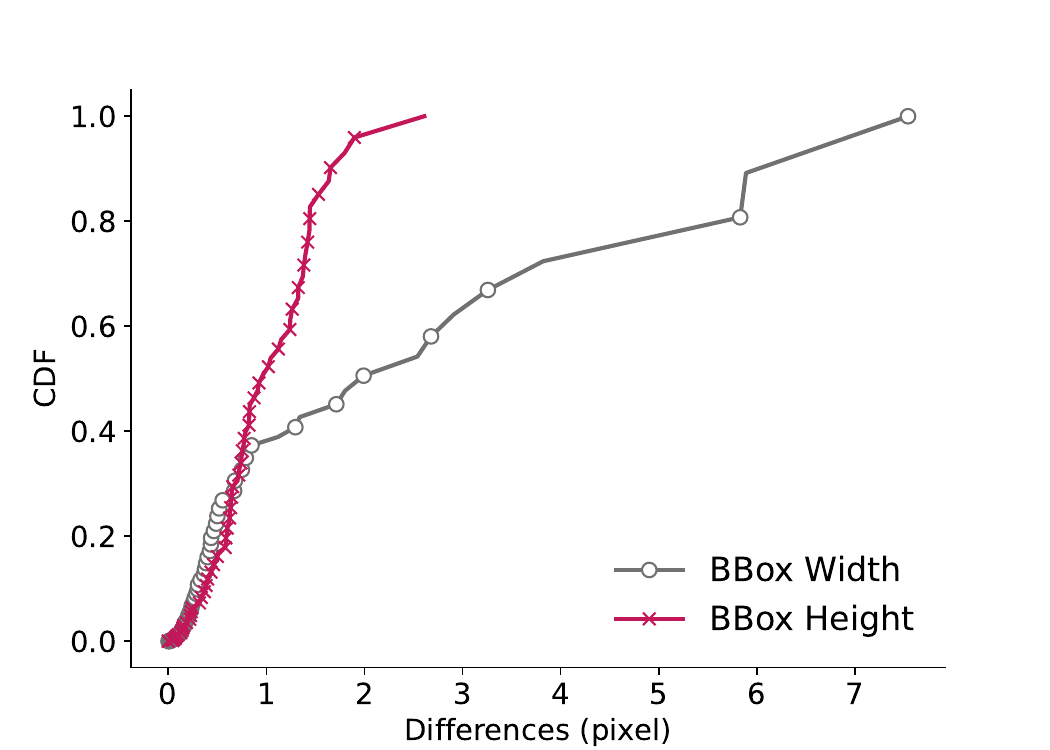}} \hspace{.5em}
        \subcaptionbox{Depth estimation using stereo vision.}
        {\includegraphics[width=0.28\textwidth]{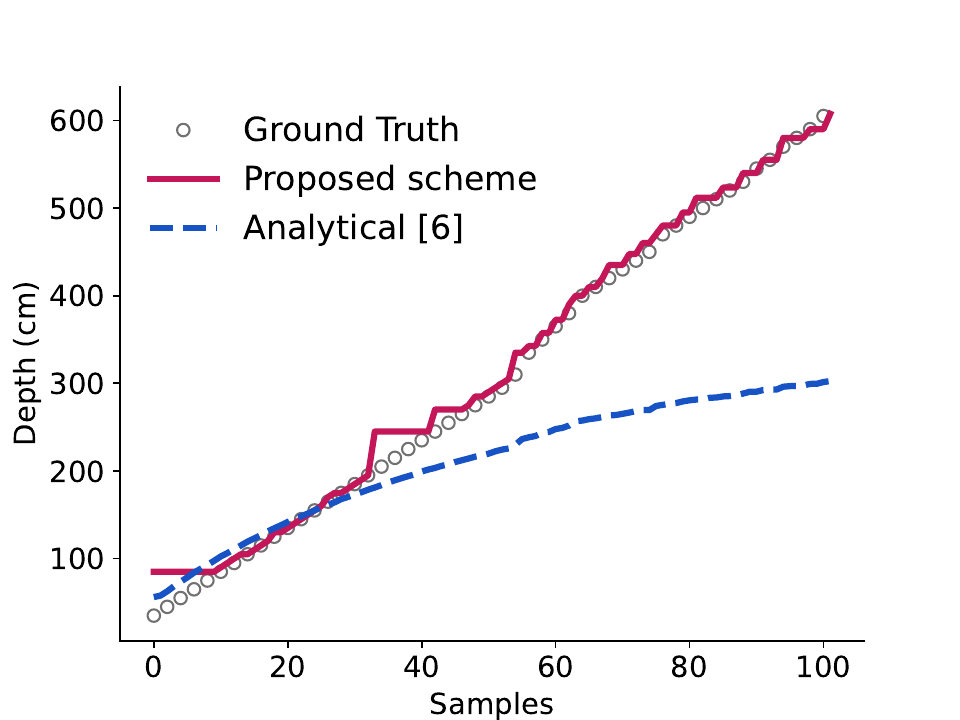}} \hspace{.5em}
        \subcaptionbox{Size estimation using stereo vision.}
        {\includegraphics[width=0.28\textwidth]{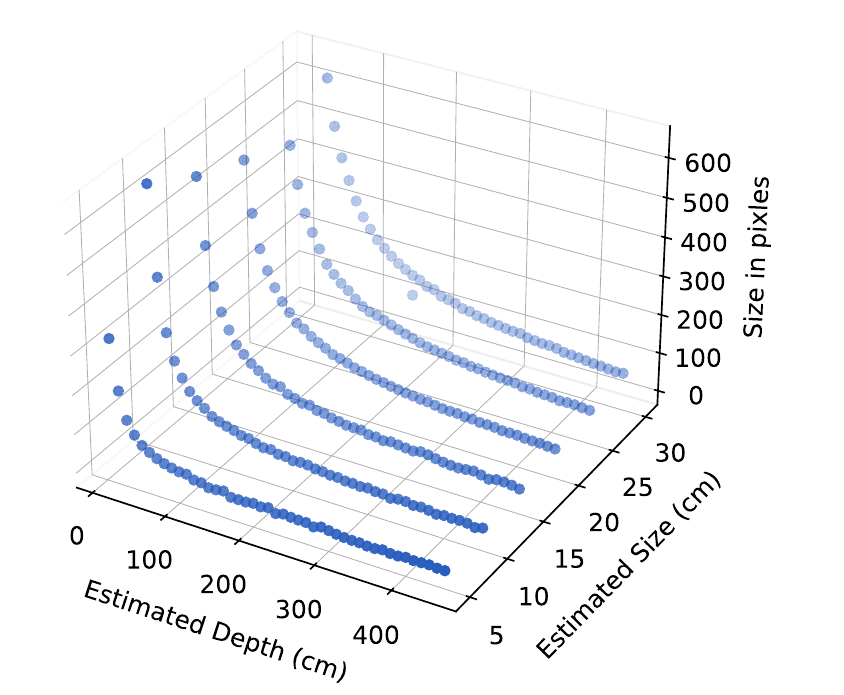}} 
        \caption{Evaluation of the proposed stereo-vision framework for obstacle avoidance.} \label{fig:results}
    \end{figure*}

In the first step, objects of interest are detected using an object detection model. Though we achieved close to 100\% accuracy for detecting obstacles, this was performed using an off-the-shelf YOLOv8 model fine-tuned on the relevant scenario. The detected objects in the form of object class and the bounding boxes are used in the depth estimation process. Fig. \ref{fig:results}a demonstrates and analyzes that the bounding box sizes are equal for the same object in both stereo images. For both the width and height of the bounding boxes, the difference is no more than 3 and 8 pixels respectively.

\begin{figure*}[!h]
    \centering
    \subcaptionbox{Obstacle detection with depth and size estimation.} 
        {\includegraphics[width=5.2cm, height=3.8cm]{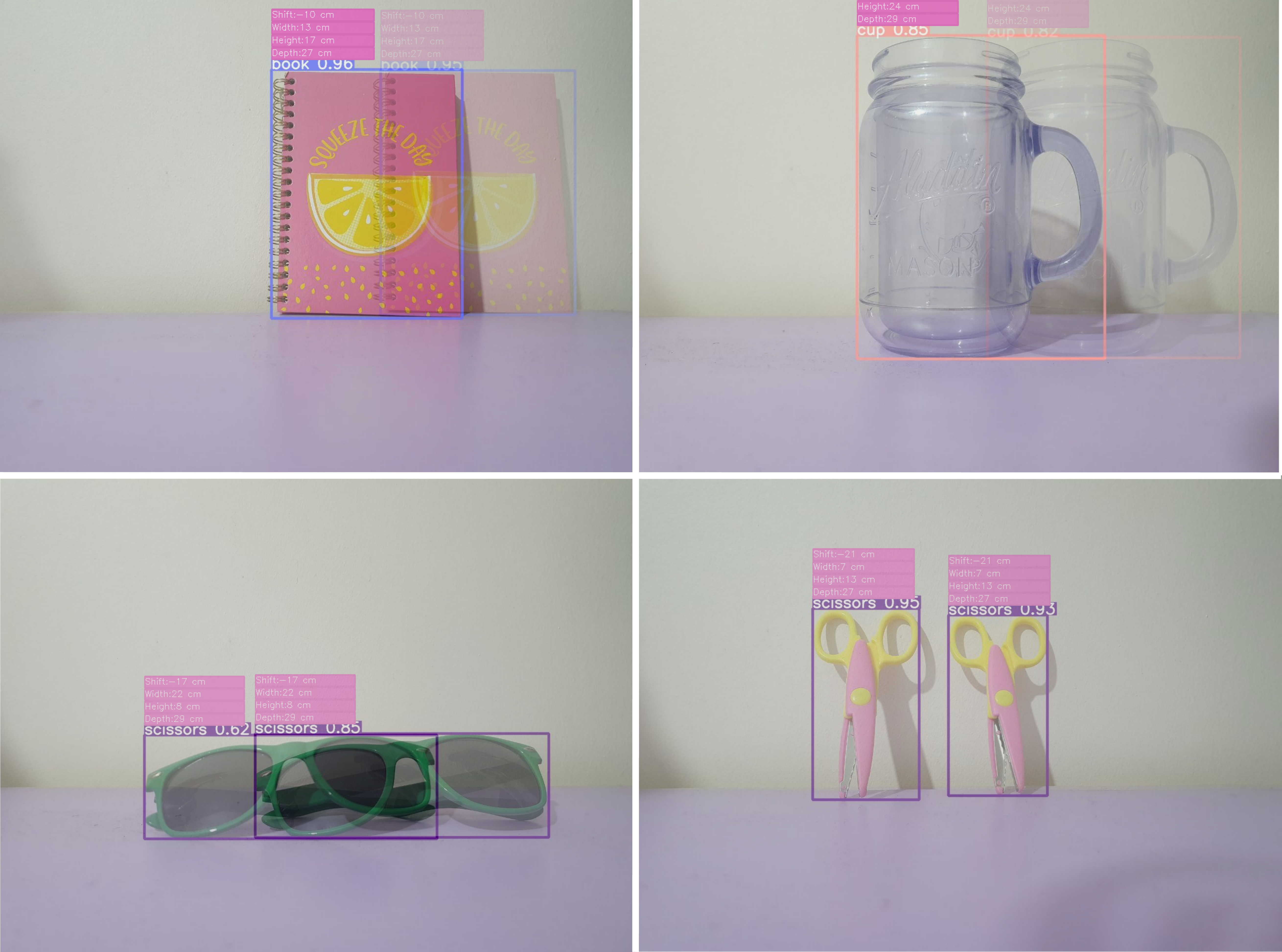}}
        \hspace{0.5em}
        \subcaptionbox{Obstacle avoidance using SLAM (with LiDAR).}
        {\includegraphics[width=5.2cm, height=3.8cm]{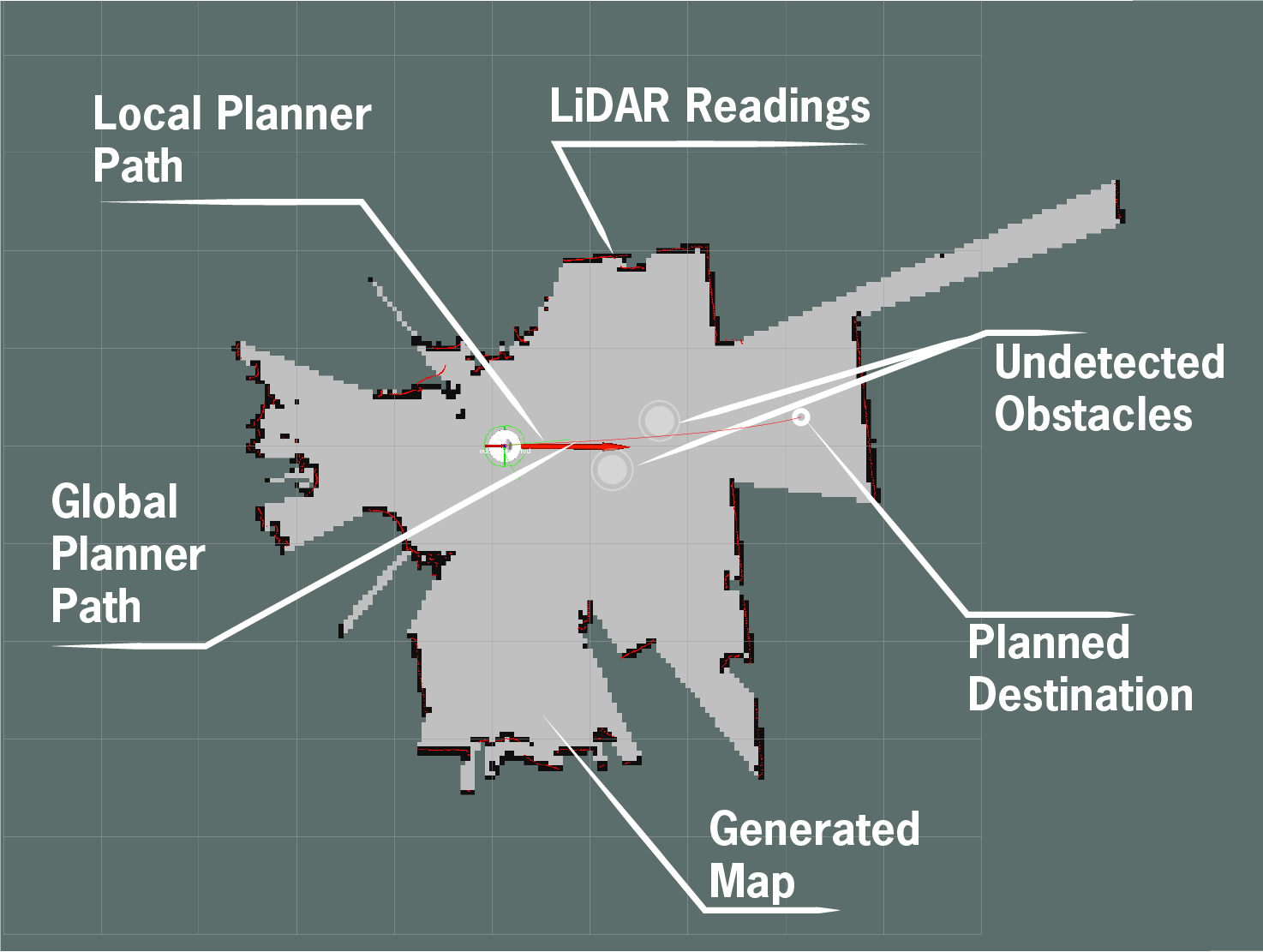}} 
        \hspace{.5em}
        \subcaptionbox{Obstacle avoidance using SLAM (with proposed scheme).} 
        {\includegraphics[width=5.2cm, height=3.8cm]{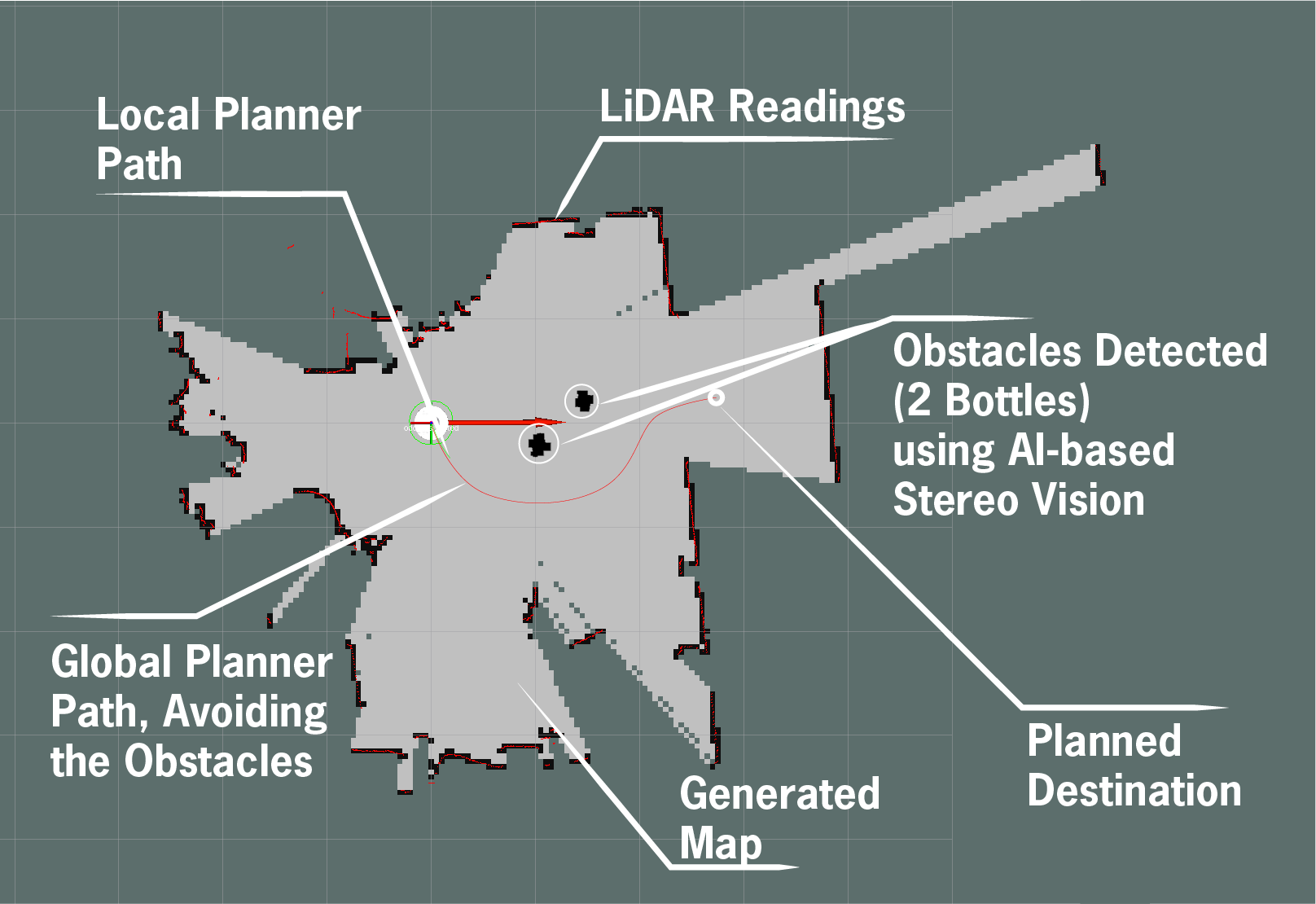}} 
        
        \caption{Integration of the proposed method in SLAM for object avoidance and navigation.} \label{fig:slam_intgration}
    \end{figure*}

For depth estimation, we first computed disparity values using Eq. \ref{eq:disparity} and then used the XGBoost regression model with a polynomial degree of $5$ to achieve the best results. Fig. \ref{fig:results}b shows the regression model results against the ground truth and the analytical model \cite{Strbac2020}. It can be quickly observed that unlike the regression model (proposed), the analytical model produces correct results only when the object is no farther than 40m. The results significantly degrade when the actual object distance increases. On the other hand, the proposed scheme produces more reliable and accurate results for all distances. The regression model is further evaluated over two common regression metrics \cite{khan2023lcdnet} i.e., mean absolute error (MAE) (Eq. \ref{eq:mae}) and mean squared error (MSE) (Eq. \ref{eq:mse}).

\begin{equation} \label{eq:mae}
    MAE = \frac{1}{n} \sum_{1}^{n}{(P(x_i) - y_i)}
\end{equation}

\begin{equation} \label{eq:mse}
    MSE = \frac{1}{n} \sum_{1}^{n}{(P(x_i) - y_i)^2}
\end{equation}
Where, $n$ is the number of data points, $P(x_i)$ is the predicted depth using the polynomial regression for the disparity $x_i$, and $y_i$ is the actual real-world depth corresponding to the disparity.

% \begin{table}[!h]
% \centering
% \renewcommand{\arraystretch}{1.0}
% \setlength{\tabcolsep}{10pt}
% \caption{Accuracy of the depth estimation results using the analytical method and the proposed scheme.} \label{tab:results}
% \begin{tabular}{|c|c|c|c|c|} \hline
                
% \multirow{2}{*}{Method} &\multicolumn{2}{c|}{Depth}  &\multicolumn{2}{c|}{Size} \\ \cline{2-5}
%                  & MSE  &MAE    & MSE  &MAE \\ \hline
% Analytical Model & 0    &0      & 0    &0 \\
% Proposed Scheme  & 21.2    &2.6      & 0    &0 \\
% \hline
% \end{tabular}
% \end{table}

Results show that the proposed scheme achieves an MAE value as low as $2.6$ and MSE of $21.2$ for depth estimation. After the depth calculation, the estimated depth information, and the bounding box dimension (width/height in pixels) as input, we trained an XGBoost regression model to estimate the respective real-world dimension of the object. Fig. \ref{fig:results}c presents the size estimation results for different depth values. Fig. \ref{fig:slam_intgration}a shows the obstacle detection results for four different objects.  While detection with SLAM (using LiDAR) was poor (\ref{fig:slam_intgration}b), the proposed scheme achieves better detection (\ref{fig:slam_intgration}c) performance and accuracy.

%===========================================
\section{Conclusion}
%===========================================
We proposed a novel approach for mobile robots to detect obstacles, and estimate the depth (horizontal distance) and dimensions (width and height) of the object in the real world using stereo vision. The proposed scheme uses only the predicted bounding box (using any off-the-shelf object detection model) without relying on other sensors (e.g., LiDAR). Highly accurate results are achieved with simple regression models using XGBoost. \red{Though the proposed method can detect non-tangible objects that can not be easily detected by LiDAR, it inherits all the limitations of vision object detection (e.g., occlusions, poor visibility, distance, size of the object, etc.).}

% use section* for acknowledgment
\section*{Acknowledgment}
This publication was made possible by the PDRA award PDRA7-0606-21012 from the Qatar National Research Fund (a member of The Qatar Foundation). The statements made herein are solely the responsibility of the authors. The publication of this article was funded by Qatar National Library.

\bibliographystyle{ieeetr}
\bibliography{biblio}
% that's all folks
\end{document}